\documentclass[10pt,twocolumn,letterpaper]{article}

\usepackage{cvpr}              

\usepackage{graphicx}
\usepackage{amsmath}
\usepackage{amssymb}
\usepackage{booktabs}
\usepackage{multirow}
\usepackage{makecell}
\usepackage[accsupp]{axessibility}
\usepackage[dvipsnames]{xcolor}

\usepackage[pagebackref,breaklinks,colorlinks]{hyperref}

\usepackage[capitalize]{cleveref}
\crefname{section}{Sec.}{Secs.}
\Crefname{section}{Section}{Sections}
\Crefname{table}{Table}{Tables}
\crefname{table}{Tab.}{Tabs.}

\newcommand{\methodname}{GateHUB}

\def\cvprPaperID{3084} 
\def\confName{CVPR}
\def\confYear{2022}

\begin{document}

\title{GateHUB: Gated History Unit with Background Suppression for Online Action Detection}




\author{Junwen Chen$^\star$$^\ddag$ ~~~~ Gaurav Mittal$^\star$$^\dag$ ~~~~ Ye Yu$^\dag$ ~~~~ Yu Kong$^\ddag$ ~~~~  Mei Chen$^\dag$\\
$^\dagger$Microsoft ~~~~~~~~~~~~~~~~~~~~~~~~ $^\ddag$Rochester Institute of Technology\\
{\tt\small \{gaurav.mittal, yu.ye, mei.chen\}@microsoft.com} ~~~
{\tt\small \{jc1088,yu.kong\}@rit.edu}
}
\maketitle

\begin{abstract}
\vspace{-0.5em}
Online action detection is the task of predicting the action as soon as it happens in a streaming video. A major challenge is that the model does not have access to the future and has to solely rely on the history, i.e., the frames observed so far, to make predictions.
It is therefore important to accentuate parts of the history that are more informative to the prediction of the current frame.
We present GateHUB, \textbf{Gate}d \textbf{H}istory \textbf{U}nit with \textbf{B}ackground Suppression, that comprises a novel position-guided gated cross-attention mechanism to enhance or suppress parts of the history as per how informative they are for current frame prediction. 
GateHUB further proposes Future-augmented History~(FaH) to make history features more informative by using subsequently observed frames when available.
In a single unified framework, GateHUB integrates the transformer's ability of long-range temporal modeling and the recurrent model's capacity to selectively encode relevant information. 
GateHUB also introduces a background suppression objective to further mitigate false positive background frames that closely resemble the action frames.
Extensive validation on three benchmark datasets, THUMOS, TVSeries, and HDD, demonstrates that 
GateHUB significantly outperforms all existing methods and is also more efficient than the existing best work.
Furthermore, a flow-free version of GateHUB is able to achieve higher or close accuracy at 2.8$\times$ higher frame rate compared to all existing methods that require both RGB and optical flow information for prediction.

\end{abstract}

\vspace{-1.5em}
\section{Introduction}
\let\thefootnote\relax\footnote{$^\star$Authors with equal contribution.}
\let\thefootnote\relax\footnote{This work was done as Junwen Chen’s internship project at Microsoft.}
\label{sec:intro}
\begin{figure}[t]
\begin{center}
    \includegraphics[width=\columnwidth]{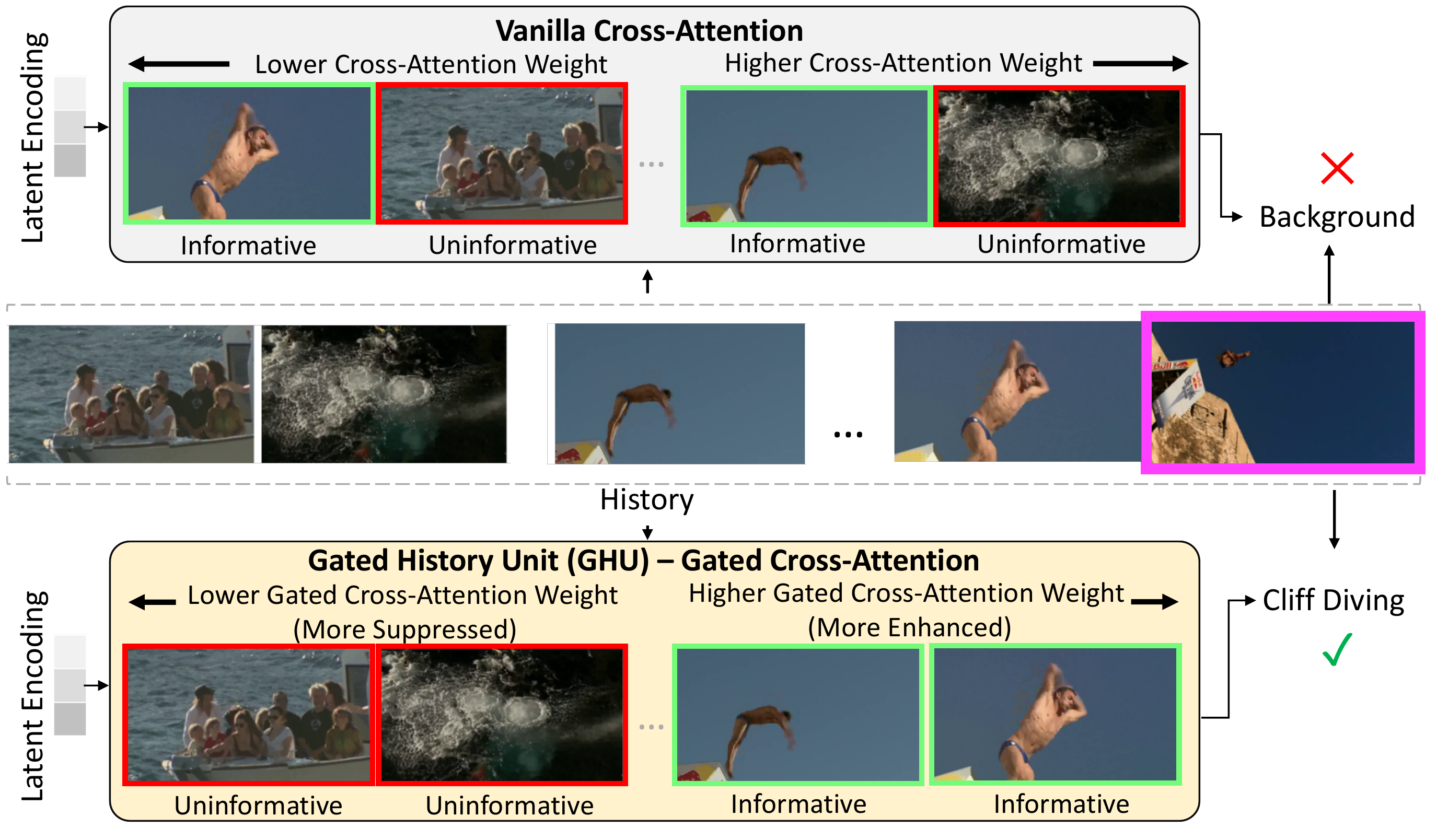}
\end{center}
\vspace{-1.5em}
\caption{We show an example video stream (middle row) where the current frame~({\color{magenta}magenta}) contains \textit{Cliff Diving} action. Weights from vanilla cross-attention~(top row) do not correlate with how informative each history frame is to current frame prediction, leading to incorrect prediction of \textit{Background}. Our novel Gated History Unit~(GHU) (bottom row) calibrates cross-attention weights using gating scores to enhance history frames that are informative to current frame prediction ({\color{green}green}) and suppress uninformative ones ({\color{red}red}), leading to accurate prediction of \textit{Cliff Diving}.
}
\label{fig:teaser}
\vspace{-1.5em} 
\end{figure}
Online action detection is the task to predict actions in a streaming video as they unfold~\cite{de2016online}. It is critical to applications including autonomous driving, public safety, virtual and augmented reality. Unlike action detection in the offline setting, where the entire untrimmed video is observable at any given moment, a major challenge for online action detection is that the predictions are solely based on observations of history without access to video frames in the future. The model needs to build a causal reasoning of the present in correlation to what happened hitherto, and as efficiently as possible for the online setting.

Prior work for online action detection~\cite{xu2019temporal, eun2020learning, eun2021temporal, gao2020woad,qu2020lap,zhao2020privileged} include recurrent-based LSTMs~\cite{hochreiter1997long} and GRUs~\cite{cho2014learning} that are prone to forgetting informative history as sequential frame processing is ineffective in preserving long-range interactions. Emerging methods~\cite{wang2021oadtr,xu2021long} employ transformers~\cite{vaswani2017attention} to mitigate this by encoding sequential frames in parallel via self-attention. Some improve model efficiency by using cross-attention~\cite{xu2021long, jaegle2021perceiverio} to compress the video sequence into a fixed-sized latent encoding for prediction. 

Fig.~\ref{fig:teaser} shows an example video stream (middle row) where the latest~(current) frame contains \textit{Cliff Diving} action. It is worth noting that, as commonly observed in video sequences, not every history frame is informative for current frame prediction (\eg frames showing people cheering or camera panning in Fig.~\ref{fig:teaser}).
Existing transformer-based approaches~\cite{xu2021long} use vanilla cross-attention to learn attention weights for history frames that determine their contribution to the current frame prediction. Such attention weights do not correlate with how informative each history frame is to current frame prediction.
As shown in Fig.~\ref{fig:teaser}~(top row), when history frames are ordered from lower to higher cross-attention weights for vanilla cross-attention, frames that are informative for current frame prediction may have lower weights while uninformative frames may have higher weights, leading to incorrect current frame prediction. Another common challenge for existing methods is false positive prediction for background frames that closely resemble action frames~(\eg pre-shot routine before golf swing). 
Existing methods also do not leverage that although future frames are not available for the current frame prediction, subsequently observed frames that are future to the history can be leveraged to enhance history encoding, which in return improves current frame prediction.

To address the above limitations, we propose \methodname, \textbf{Gate}d \textbf{H}istory \textbf{U}nit with \textbf{B}ackground suppression. \methodname~comprises a novel Gated History Unit~(GHU), a position-guided gated cross-attention module that enhances informative history while suppressing uninformative frames via gated cross-attention (as shown in Fig.~\ref{fig:teaser}, bottom row). GHU enables \methodname~to encode more informative history into the latent encoding to better predict for current frame. GHU combines the benefit of an LSTM-inspired gating mechanism to filter uninformative history with the transformer’s ability to effectively learn from long sequences.

\methodname~leverages \textit{future frames for history} by introducing Future-augmented History~(FaH). FaH extracts features for a history frame using its future, \ie the subsequently \textit{observed} frames. This makes a history frame aware of its future and helps it to be more informative for current frame prediction. To tackle the common false positives in prior art, \methodname~proposes a novel background suppression objective that has different treatments for low-confident action and background predictions. These novel approaches enable \methodname~to outperform all existing methods on common benchmark datasets THUMOS~\cite{THUMOS14}, TVseries~\cite{de2016online}, and HDD~\cite{ramanishka2018toward}.  Keeping model efficiency in mind for the online setting, we also validate that \methodname~is more efficient than the existing best method~\cite{xu2021long} while being more accurate. Moreover, our proposed optical flow-free variant is $2.8\times$ faster than all existing methods that require both RGB and optical flow data with higher or close accuracy.
To summarize, our main contributions are:
\begin{enumerate}
	\item Gated History Unit~(GHU), a novel position-guided gated cross-attention that explicitly enhances or suppresses parts of video history as per how informative they are to predicting action for the current frame.
	\item Future-augmented History~(FaH) to extract features for a history frame using its subsequently observed frames to enhance history encoding. 
	\item A background suppression objective to mitigate the false positive prediction of background frames that closely resemble the action frames.
	\item \methodname~is more accurate than all existing methods and is also more efficient than the existing best work. Moreover, our proposed optical flow-free model is $2.8\times$ faster compared to all existing methods that require both RGB and optical flow information while achieving higher or close accuracy.
\end{enumerate}
\section{Related Work}
\begin{figure*}[t]
\begin{center}
\includegraphics[width=\textwidth]{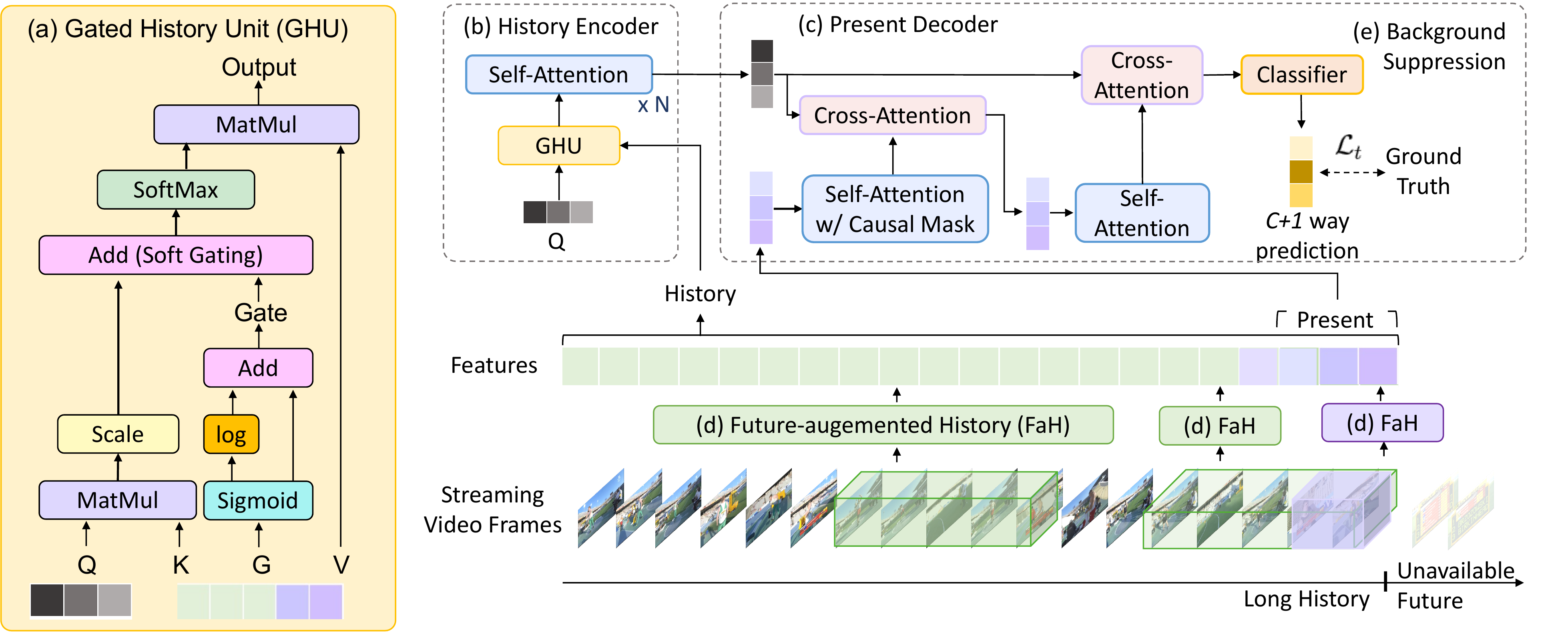}
\end{center}
\vspace{-1.5em}
\caption{{\bf Model Overview.} \methodname~comprises a novel Gated History Unit~(GHU)~(a) as part of History Encoder~(b) to explicitly enhance or suppress history frames, \ie streaming video frames observed so far, as per how informative they are to current frame prediction. GHU encodes them by cross-attending with a latent encoding~(Q). \methodname~uses Future-augmented History features~(FaH)~(d) to encode each history frame using $t_f$ subsequently observed future frames. The Present Decoder~(c) correlates with history by cross-attending the encoded history with the present, \ie, a small set of most recent frames, to make current frame prediction. We subject the prediction to a background suppression loss~(e) to reduce false positives by effectively separating action frames from closely resembling background frames.} 
\label{fig:gatehub_model}
\vspace{-1em}
\end{figure*}

{\noindent \bf Online Action Detection.} 
Previous methods for online action detection include use 3D ConvNet~\cite{de2016online}, reinforcement learning~\cite{gao2017red}, recurrent networks~\cite{xu2019temporal, eun2020learning, qu2020lap, gao2020woad,zhao2020privileged,qu2020lap} and more recently, transformers~\cite{wang2021oadtr, xu2021long}. The primary challenge in leveraging history is that for long untrimmed videos, its length becomes intractably long over time. To make it computationally feasible, some ~\cite{eun2020learning, wang2021oadtr, gao2020woad,qu2020lap} make the online prediction conditioned only on the most recent frames spanning less than a minute. This way the history beyond 
this duration 
that might be informative to current frame predictions is left unused. 
TRN~\cite{xu2019temporal} mitigates this by the hidden state in LSTMs~\cite{hochreiter1997long} to memorize the entire history during inference.
But LSTM limits its ability to model long-range temporal interactions. More recently, \cite{xu2021long} proposes to scale transformers to the history spanning longer duration.
However, not every history frame is informative and useful. \cite{xu2021long} lacks the forgetting mechanism of LSTM to filter uninformative history which causes it to encode uninformative history into the encoding leading to incorrect predictions. Our Gated History Unit~(GHU) and Future-augmented History~(FaH) combine the benefits of LSTM's selective encoding and transformer's long range modeling to leverage long-duration history more informatively to outperform all previous methods. 
{\noindent \bf Transformers for Video Understanding.} Transformers can achieve superior performance on video understanding tasks by effectively modeling the spatio-temporal context via attention. Most of the previous transformer-based methods~\cite{bertasius2021space,arnab2021vivit,fan2021multiscale, neimark2021video} focus on action recognition in trimmed videos~\cite{carreira2017quo}~(videos spanning few seconds) due to the quadratic complexity \wrt video length. Untrimmed videos have a longer duration from a few minutes to hours and contain frames with irrelevant actions~(labeled as background). Temporal action localization (TAL)~\cite{shou2016temporal,xu2017r,gao2017turn,shou2017cdc,Zhao_2017_ICCV,buch2017sst,liu2019multi,lin2019bmn,zhu2021enriching,zhang2021temporal} and temporal action proposal generation (TAP)~\cite{lin2018bsn,lin2019bmn,tan2021relaxed} are two fundamental tasks in untrimmed video understanding. AGT\cite{nawhal2021activity} proposes activity graph transformer for TAL based on DETR~\cite{carion2020end}. TAPG\cite{wang2021temporal} applies transformer to predict the activity boundary for TAP. However, unlike TAL and TAP which are both offline tasks having access to the entire video, online action detection does not have access to the future and requires causal understanding from history to present. We follow the existing transformer-based streaming tasks\cite{girdhar2021anticipative, chen2021developing, xu2021long} and apply a causal mask to address online action detection.
{\noindent \bf Long Sequence Modeling.} To model long input sequences, recent work~\cite{dosovitskiy2020image} proposes to reduce complexity by factorizing~\cite{touvron2021training} or subsampling the inputs~\cite{chen2020generative}. Another group of work focuses on modifying the internal dense self-attention module to boost the efficiency~\cite{beltagy2020longformer,wang2020linformer}. More recently, Perceiver~\cite{jaegle2021perceiver} and PerceiverIO~\cite{jaegle2021perceiverio} propose to cross-attend long-range inputs to a small fixed-sized latent encoding, adding further flexibility in terms of input and reducing the computational complexity. However, unlike our GHU, PerceiverIO lacks an explicit mechanism to enhance/suppress history frames making it sub-optimal for online action detection. Our method uses LSTM-inspired gating to calibrate cross-attention to enhance/suppress history frames per their informative-ness while employing transformers to learn from long history sequences effectively.
\section{Methodology}

Given a streaming video sequence
$\mathbf{h} = [h_t]^0_{t=-T+1}$,
our task is to identify \textit{if} and \textit{what} action $y_0\in\left \{0,1,..., C \right \}$ occurs at the current frame $h_0$. We have a total of $C$ action classes and label 0 for background frames with no action. Since future frames $h_1, h_2, ...$, are NOT accessible, the model makes the $C+1$-way prediction for the current frame based on the recent $T$ frames, $[h_t]^0_{t=-T+1}$, 
observed up until the current frame. While $T$ may be large in an untrimmed video stream, as shown in the top row of Fig.~\ref{fig:teaser}, all frames observed in past history $[h_t]^{-1}_{t=-T+1}$ may not be  equally informative to the prediction for the current frame.

\subsection{Gated History Unit based History Encoder}
To make the $C+1$-way prediction accurately for current frame $h_0$ based on $T$ history frames, $\mathbf{h} = [h_t]^0_{t=-T+1}$, we employ transformers to first encode the video sequence history and then associate the current frame with the encoding for prediction. Inspired by the recently introduced PerceiverIO~\cite{jaegle2021perceiverio}, our method consists of a History Encoder (Fig.~\ref{fig:gatehub_model}b) that uses cross-attention to project the variable length history to a fixed-length learned latent encoding. Using cross-attention is more efficient than using self-attention because its computational complexity is quadratic \wrt latent encoding size instead of the video sequence length which is typically orders of magnitude larger. This is crucial to developing a model for the online setting. However, as shown in Fig.~\ref{fig:teaser}, vanilla cross-attention, as used in PerceiverIO and LSTR~\cite{xu2021long}, fails to learn attention weights for history frames that correlate with how informative each history frame is for $h_0$ prediction. 
We therefore introduce a novel Gated History Unit~(GHU) (Fig.~\ref{fig:gatehub_model}a) that has a position-guided gated cross-attention mechanism which learns a set of gating scores $G$ to calibrate the attention weights to effectively enhance or suppress history frames based on how informative they are to current frame prediction. 

Specifically, given $\mathbf{h} = [h_t]^0_{t=-T+1}$ 
as the streaming sequence of $T$ history frames ending at current frame $h_0$, we encode $\mathbf{h}$ with a feature extraction backbone, $u$, followed by a linear encoding layer $\mathbf{E}$. We then subject the output to a learnable position encoding, $\mathbf{E_{pos}}$, relative to the current frame, $h_0$, to give  $\mathbf{z^h} = u(\mathbf{h})\mathbf{E} + \mathbf{E_{pos}}$
where $u(\mathbf{h}) \in \mathbb{R}^{T \times M}$, $\mathbf{E} \in \mathbb{R}^{M \times D}$, $\mathbf{z^h} \in \mathbb{R}^{T \times D}$ and $\mathbf{E_{pos}} \in \mathbb{R}^{T \times D}$. $M$ and $D$ denote the dimensions of extracted features and post-linear encoding features, respectively. We also define a learnable latent query encoding, $\mathbf{q} \in \mathbb{R}^{L \times D}$, that we cross-attend with $\mathbf{h}$. Following the standard multi-headed cross-attention setup~\cite{jaegle2021perceiver, jaegle2021perceiverio}, let $N_{heads}$ be the number of heads in GHU such that $Q_i = \mathbf{q} \mathbf{W}_i^q$, $K_i = \mathbf{z^h} \mathbf{W}_i^k$, $V_i = \mathbf{z^h} \mathbf{W}_i^v$ be the queries, keys and values, respectively, for each head $i \in \{1, \ldots, N_{heads}\}$~(Fig.~\ref{fig:gatehub_model}a) where projection matrices $\mathbf{W}_i^q, \mathbf{W}_i^k \in \mathbb{R}^{D \times d_k}$ and $\mathbf{W}_i^v \in \mathbb{R}^{D \times d_v}$. We assign $d_k = d_v = D/N_{heads}$ in our set up~\cite{vaswani2017attention}. Next, we obtain the position-guided gating scores, $G$, for $\mathbf{h}$ as,
\begin{align}
\small
    \mathbf{z^g} &= \sigma(\mathbf{z^h}\mathbf{W}^g)  \label{eqn:gate_sigmoid} \\
    G &= \log(\mathbf{z^g}) + \mathbf{z^g} \label{eqn:gate_val}
\end{align}
where $\mathbf{W}^g \in \mathbb{R}^{D \times 1}$ is the matrix projecting each history frame to a scalar. $\mathbf{z^g} \in \mathbb{R}^{T \times 1}$ is a sequence of scalars for the history frames $\mathbf{h}$ after applying sigmoid $\sigma$. $G \in \mathbb{R}^{T \times 1}$ is the gating score sequence for history frames in GHU. By using $\mathbf{z^h}$ which already contains the position encoding, the gates are guided by the relative position of the history frame to the current frame $h_0$. As also shown in Fig.~\ref{fig:gatehub_model}a, we now compute the gated cross-attention for each head, $GHU_i$, as,
\begin{equation}
\label{eqn:ghu}
\small
    GHU_i = \text{Softmax}\left(\frac{Q_iK_i^T}{\sqrt{d_k}} + G \right) V_i
\end{equation}
and multi-headed gated cross-attention defined as,
\begin{equation}
\small
\text{MultiHeadGHU}(Q, K, V, G) = \text{Concat}([GHU_i]_{i=0}^{N_{heads}}) \mathbf{W^o}
\end{equation}
where $\mathbf{W}^o \in \mathbb{R}^{D \times D}$ re-projects the attention output to $D$ dimension.  It is possible to define $G$ separately for each head but in our method, we find sharing $G$ across all heads to perform better~(Sec.~\ref{sec:ablation}). From Eqn.~\ref{eqn:gate_sigmoid} and \ref{eqn:gate_val}, we can observe that each scalar in $\mathbf{z^g}$ lies in $[0, 1]$ due to sigmoid which implies that each gating score in $G$ lies in $[-\inf, 1]$. This enables the softmax function in Eqn.~\ref{eqn:ghu} to calibrate the attention weight for each history frame by a factor in $[0, e]$ such that a factor in $[0, 1)$ suppresses a given history frame and a factor in $(1, e]$ enhances a given history frame. This provides an explicit ability to GHU to learn to calibrate the attention weight of a history frame based on how informative the history frame is for prediction of $h_0$. 
Unlike previous methods with relative position bias~\cite{liu2021swin,dai2019transformer}, $G$ is input-dependent and learns based on the history frame and its position \wrt current frame. This enables GHU to assess how informative each history frame is based on its feature representation and relative position from the current frame $h_0$. We feed the output of GHU to a series of $N$ self-attention layers to obtain the final history encoding~(Fig.~\ref{fig:gatehub_model}b).

\subsection{Hindsight is 2020: Future-augmented History}
Existing methods~\cite{wang2021oadtr,xu2019temporal,xu2021long, gao2020woad, eun2020learning} extract features for each frame by feed-forwarding the frame and optionally, a small set of past consecutive frames through pretrained networks like TSN~\cite{wang2016temporal} and I3D~\cite{carreira2017quo}. It is worth noting that although for current frame prediction its future is not available, for the history frames their \textit{future} is accessible and this \textit{hindsight} can potentially improve the encoding of history for current frame prediction. Existing methods do not have a mechanism to leverage this.
This inspires us to propose a novel feature extraction scheme, Future-augmented History~(FaH), where we aggregate observed future information into the features of a history frame to make it aware of its so far observable future. 
Fig.~\ref{fig:gatehub_model}d illustrates the FaH feature extraction process. For a history frame $h_t$ and a feature extraction backbone $u$, when $t_f$ \textit{future history frames} for $h_t$ can be observed, FaH extracts features for $h_t$ using a set of frames $[h_i]_{i=t}^{t+t_f}$ (\ie history frame itself and its subsequently observed $t_f$ future frames). Otherwise, FaH extracts features for $h_t$ using a set of frames $[h_i]_{i=t - t_{ps}}^t$ (\ie history frame itself and its past $t_{ps}$ frames),
\begin{flalign}
\label{eqn:fah}
\small
u(h_t)=\left\{\begin{matrix}
u([h_i]_{i=t - t_{ps}}^{t}) & \text{if}\: t>-t_f \\ 
u([h_i]_{i=t}^{t+t_f}) & \text{if}\: t <= -t_f \\ 
\end{matrix}\right.
\end{flalign}
At each new time step with one more new frame getting observed, FaH will feed-forward through $u$ twice to extract features for (1) the new frame using $[h_i]_{i=-t_{ps}}^0$ frames and (2) $h_{-t_f}$ that is now eligible to aggregate future information using $[h_i]_{i=-t_f}^0$ frames~(as shown in Fig.~\ref{fig:gatehub_model}d purple and green cuboid respectively). Thus, FaH has the same time complexity as existing feature extraction methods. FaH does not trivially incorporate all available subsequently observed frames. Instead, it encodes only from a set of future frames that are the most relevant to a history frame (as we empirically explain later in Section~\ref{sec:ablation}).
\subsection{Present Decoder}
In order to correlate the present with history to make current frame prediction, we sample a subset of $t_{pr}$ most recent history frames $[h_t]_{t=-t_{pr}-1}^0$ to model the present~(\ie the most immediate context) for $h_0$ using the Present Decoder~(Fig.~\ref{fig:gatehub_model}c). After extracting the features via FaH, we apply a learnable position encoding, $\mathbf{E^{pr}_{pos}}$, to each of the $t_{pr}$ frame features and subject them to a multi-headed self-attention with a causal mask. The causal mask limits the influence of only the preceding frames on a given frame. We then cross-attend the output from self-attention with the history encoding from the History Encoder. Inspired by Perceiver~\cite{jaegle2021perceiver}, we repeat this process twice and the self-attention does not need a causal mask the second time. Finally, we feed the output corresponding to each of $t_{pr}$ frames to the classifier layer for prediction.

\subsection{Background Suppression Objective}
\label{sec:bg_suppression}
Existing online action detection methods~\cite{wang2021oadtr,xu2019temporal,xu2021long, gao2020woad, eun2020learning} apply standard cross entropy loss for $C+1$-way multi-label per-frame prediction. Standard cross entropy loss does not consider that the ``no action'' background class does not belong to any specific action distribution and is semantically different from the $C$ action classes. This is because background frames can be anything from completely blank at the beginning of a video to closely resemble action frames without actually being action frames (\eg, aiming before making a billiards shot). The latter is a common cause for false positives in online action detection. In addition to the complex distribution of background frames, untrimmed videos suffer from a sharp data imbalance where background frames significantly outnumber action frames. 

To tackle these challenges, we design a novel background suppression objective that applies separate emphasis on low-confident action and background predictions during training to increase the margin between action and background frames~(Fig.~\ref{fig:gatehub_model}e). Inspired by focal loss~\cite{lin2017focal}, our objective function, $\mathcal{L}_t$ for frame $h_t$ is defined as,
\begin{align}
\small
\mathcal{L}_t=\left\{\begin{matrix}
-  y^0_t (1-p^0_t)^{\gamma_b}\log(p^0_t) & \text{if}\: y^0_t=1\\ 
 - \Sigma_{i=1}^C y^i_t (1-p^i_t)^{\gamma_a}\log(p^i_t) & \text{otherwise} \\ 
\end{matrix}\right.
\end{align}
where $\gamma_a, \gamma_b>0$ enables low-confident samples to contribute more to the overall loss forcing the model to put more emphasis on correctly predicting these samples. Unlike original focal loss~\cite{lin2017focal}, our background suppression objective specializes for online action detection by applying separate $\gamma$ to action classes and background. This separation is necessary to distinguish the action classes that have a more constrained distribution from the background class whose distribution is more complex and unconstrained. Our objective is the first attempt in online action detection to put separate emphasis on low-confident hard action and background predictions.

\subsection{Flow-free Online Action Detection}
Existing methods~\cite{xu2019temporal, wang2021oadtr, eun2020learning} for online action detection use optical flow in addition to RGB to capture fine-grained motion among frames. Computing optical flow takes much more time than feature extraction or model inference, and can be unrealistic for time-critical applications such as autonomous driving. This motivates us to develop an optical flow-free version of \methodname~that is able to achieve higher or close accuracy compared to existing methods without time-consuming optical flow estimation.
To capture motion without optical flow using only RGB frames, we leverage multiple temporal resolutions 
using a spatio-temporal backbone such as TimeSformer~\cite{bertasius2021space}. We extract two feature vectors for a frame $h_t$ by encoding a frame sequence sampled at a higher frame rate spanning a smaller time duration and another frame sequence sampled at a lower frame rate spanning a larger time duration. Similar to the setup using RGB and optical flow features, we concatenate the two feature vectors before feeding them to \methodname.
\section{Experiments}

\subsection{Datasets}
Following existing online action detection work~\cite{wang2021oadtr, xu2019temporal, eun2020learning, gao2017red, xu2021long}, we evaluate \methodname~on three common benchmark datasets -- THUMOS'14, TVSeries, and HDD.

{\bf THUMOS’14}~\cite{THUMOS14} consists of over 20 hours of sports video and is annotated with 20 actions.
We follow prior work~\cite{wang2021oadtr, xu2019temporal} and train on the validation set~(200 untrimmed videos) and evaluate on the test set~(213 untrimmed videos). 

{\bf TVSeries}~\cite{de2016online} includes 27 episodes of 6 popular TV shows with a total duration of 16 hours. It is annotated with 30 real-world everyday actions, \eg open door, run, drink.

{\bf HDD}~(Honda Research Institute Driving Dataset)~\cite{RamanishkaCVPR2018} includes 137 driving videos with a total duration of 104 hours. Following prior work~\cite{wang2021oadtr}, we use the vehicle sensor as input signal and divide data into 100 sessions for training and 37 sessions for testing. 

\subsection{Implementation Details}
For TVSeries and THUMOS'14, following~\cite{wang2021oadtr, xu2019temporal, eun2020learning, gao2017red, xu2021long}, we resample the videos at 24 FPS~(frames per second) and then extract frames at 4 FPS for training and evaluation. The sizes of \textit{history} and \textit{present} are set to 1024 and 8 most recently observed frames, respectively, spanning durations of 256s and 2s correspondingly at 4 FPS.  
For HDD, following OadTR~\cite{wang2021oadtr}, we extract the sensor data at 3 FPS for training and evaluation. The sizes of \textit{history} and \textit{present} are 48 and 6 most recently observed frames respectively, spanning durations of 16s and 2s correspondingly at 3 FPS.

{\bf Feature Extraction.}
Following~\cite{xu2021long, wang2021oadtr}, we use mmaction2~\cite{2020mmaction2}-based two-stream TSN~\cite{wang2016temporal} pretrained on Kinetics-400~\cite{carreira2017quo} to extract frame-level RGB and optical flow features for THUMOS'14 and TVSeries. 
We concatenate the RGB and optical flow features along channel dimension before feeding to the linear encoding layer in \methodname. For HDD, we directly feed the sensor data as input to \methodname.
To fully leverage the proposed FaH, the feature extraction backbone needs to support multi-frame input. Since TSN only supports single-frame input, we explore spatio-temporal TimeSformer~\cite{bertasius2021space}~(pretrained on Kinetics-600 using $96\times4$ frame sampling) that supports multiple-frame input. 
We set the time duration for past $t_{ps}$ and future $t_f$ frames under FaH to be 1s and 2s respectively. 
We use TimeSformer to extract RGB features and use TSN-based optical flow features as TimeSformer only supports RGB. We also demonstrate FaH using RGB features from I3D~\cite{carreira2017quo} with results in the supplementary. For our flow-free version, we replace optical flow features with features obtained from an additional multi-frame input of RGB frames uniformly sampled from a duration of 2s. Please refer to supplementary for additional details.


{\bf Training.}
We train \methodname~for 10 epochs using Adam optimizer~\cite{kingma2014adam}, weight decay of $5e^{-5}$,  batch size of 50, OneCycleLR learning rate schedule of PyTorch~\cite{paszke2017automatic} with pct\_start of 0.25, $D=1024$, latent encoding size $L=16$, number of self-attention layers in History Decoder $N=2$, $N_{heads}=16$ for each attention layer and $\gamma_a=0.6, \gamma_b = 0.2$ for background suppression.


{\bf Evaluation Metrics}
We follow the protocol of per-frame mean average precision~(mAP) for THUMOS and HDD and calibrated average precision (mcAP)~\cite{de2016online} for TVSeries.

\subsection{Comparison with State-of-the-Art}
\begin{table}[h]
\vspace{-1em}
\centering
\resizebox{\columnwidth}{!}{%
\begin{tabular}{llccc}
\hline
\multirow{2}{*}{Method} & \multicolumn{2}{c}{Feature Backbone} & THUMOS14  \\  
 & \multicolumn{1}{l}{RGB} & \multicolumn{1}{c}{Optical Flow} & mAP~(\%)  \\ \hline
      FATS~\cite{kim2021temporally}  &  \multirow{8}{*}{TSN} & \multirow{8}{*}{TSN}    & 59.0        \\
      IDN~\cite{eun2020learning}   &   &   & 60.3        \\
TRN~\cite{xu2019temporal}   &    &     & 62.1         \\

PKD~\cite{zhao2020privileged} &  &  & 64.5 \\
OadTR~\cite{wang2021oadtr} &    &    & 65.2     \\
WOAD~\cite{gao2020woad}  &     &   & 67.1        \\
LSTR~\cite{xu2021long}  &     &    & 69.5      \\
\methodname~(Ours)  &      &     & \textbf{70.7}        \\ \hline
TRN~\cite{xu2019temporal}   &  \multirow{4}{*}{TimeSformer} & \multirow{4}{*}{TSN}  &   68.5 &   \\
OadTR~\cite{wang2021oadtr} &   &    &  65.5     \\
LSTR~\cite{xu2021long}  &     &      &   69.6   \\
\methodname~(Ours)  &      &      & \textbf{72.5}       \\ \hline
\end{tabular}%
}
\vspace{-0.5em}
\caption{Online action detection results on THUMOS'14 comparing \methodname~with SoTA methods on mAP~(\%) when the RGB-based features are extracted from either TSN or TimeSformer. Optical flow-based features are extracted from TSN in all settings.}
\label{tab:thumos}
\vspace{-0.75em}
\end{table}

Table~\ref{tab:thumos} compares \methodname~with existing state-of-the-art~(SoTA) online action detection methods on THUMOS'14 for two different setups, one using RGB features from TSN~\cite{wang2016temporal} and the other using RGB features from TimeSformer~\cite{bertasius2021space}. Both setups use optical flow features from TSN. WOAD~\cite{gao2020woad} uses RGB features from I3D~(equivalent to TSN). For TSN RGB features, all mAP in Table~\ref{tab:thumos} are as reported in the references. 
For TimeSformer RGB features, we use the official code for TRN, OadTR and LSTR for fair comparison.
From the table, we can observe that \methodname~outperforms all existing methods by at least $1.2\%$ when using RGB features from TSN. Moreover, \methodname~ outperforms existing methods by a larger margin of at least $2.9\%$ using RGB features from TimeSformer.
\methodname~is also the first approach to surpass $70\%$ on THUMOS'14 benchmark. 
This validates that \methodname, comprising GHU, Background Suppression and FaH to holistically leverage the long history more informatively, outperforms all SoTA on THUMOS’14.

We further compare \methodname~with SoTA on TVSeries and HDD in Table~\ref{tab:tvseries_hdd}a and~\ref{tab:tvseries_hdd}b, respectively. Following protocol, we use RGB and optical flow features from TSN for TVSeries and sensor data for HDD. All results from SoTA are as reported in the references. We can observe that \methodname~outperforms all SoTA on both TVSeries and HDD. The large improvement on HDD using sensor data validates that \methodname~is also effective on data modalities other than RGB or optical flow. 
\begin{table}[h]
\centering
\resizebox{\columnwidth}{!}{%
\begin{tabular}{cc}  
\begin{tabular}{lc}
\hline
Method  & mcAP~(\%) \\\hline
      FATS~\cite{kim2021temporally}     & 84.6       \\
      IDN~\cite{eun2020learning}        & 86.1       \\
TRN~\cite{xu2019temporal}      & 86.2        \\

PKD~\cite{zhao2020privileged}    & 86.4  \\
OadTR~\cite{wang2021oadtr}    & 87.2         \\
LSTR~\cite{xu2021long}       & 89.1      \\
\methodname~(Ours)         & \textbf{89.6}        \\ \hline
\end{tabular}%
     &  
     \begin{tabular}{lc}
\hline
Method &  mAP~(\%) \\ \hline 
CNN~\cite{de2016online}   &   22.7 \\
LSTM~\cite{ramanishka2018toward}  &  23.8 \\
RED~\cite{gao2017red}   &   27.4 \\
TRN~\cite{xu2019temporal}  &   29.2 \\
OadTR~\cite{wang2021oadtr} &   29.8 \\
\methodname~(Ours)   & \textbf{32.1} \\ \hline
\end{tabular}%
     \\
     (a) & (b)  \\
\end{tabular}
}
\vspace{-1em}
\caption{Online action detection results comparing \methodname~with state-of-the-art methods on (a) TVSeries using RGB + Optical Flow data as input on mcAP metric and (b) HDD using sensor data as input on mAP metric.}
\label{tab:tvseries_hdd}
\vspace{-1em}
\end{table}
\begin{table*}[h]
\centering
\setlength{\tabcolsep}{4pt}
\resizebox{0.9\textwidth}{!}{%
\begin{tabular}{ccc} 
\begin{tabular}{l|c}
\hline
Method & mAP~(\%)    \\ \hline
w/ GHU (Ours)           & \textbf{70.7} \\ 
w/o GHU             & 69.6 \\
w/ GHU suppress only     & 70.5    \\ 
w/ GHU enhance only      & 70.5 \\ 
w/ GHU w/o position-guidance    & 70.3 \\ 
w/ GHU per head            &  68.0\\\hline
\end{tabular} 
&
\begin{tabular}{ll}
\hline
Method &  mAP~(\%)    \\ \hline
Ours $\gamma_a>\gamma_b$       & \textbf{70.7} \\
Ours $\gamma_a<\gamma_b$            &  70.2 \\  
w/ cross-entropy       & 69.9 \\
w/ standard focal loss      & 70.2 \\ \hline
\end{tabular}%
&
     \begin{tabular}{lcc}
\hline
Method & Future Duration & mAP~(\%)    \\ \hline
w/o FaH    & -       & 71.5 \\ \hline
\multirow{4}{*}{w/ FaH}    & 0.5       & 71.1 \\
& 1s       & 72.0 \\
& 2s        &  \textbf{72.5} \\
    & 4s        & 71.4 \\ \hline
\end{tabular}%
     \\
     (a) & (b) & (c) \\
\end{tabular}
}
\vspace{-1em}
\caption{Ablation study comparing different variants of (a) Gated History Unit~(GHU), (b) background suppression objective and (c) Future-augmented History~(FAH). Ablation in (a) and (b) is conducted with RGB features from TSN and in (c) are conducted with RGB features from TimeSformer. Optical flow features from TSN are used in all settings.
}
\label{tab:ghu_bg_fah}
\vspace{-1.25em}
\end{table*}
\subsection{\methodname: Ablation Study}
\label{sec:ablation}
In this section, we conduct an ablation study to highlight the impacts of the novel components of \methodname. Unless stated otherwise, all experiments are on THUMOS'14 using RGB and optical flow features from TSN. 

{\noindent \bf Impact of Gated History Unit~(GHU).} We conduct an experiment where we test different variants of our Gated History Unit~(GHU) by removing one or more of its design elements. Table~\ref{tab:ghu_bg_fah}a summarizes the results of this experiment. In the table, `w/o GHU' refers to replacing GHU with vanilla cross-attention from Perceiver IO~\cite{jaegle2021perceiverio} and LSTR~\cite{xu2021long}, \ie, $\text{CrossAttention}(Q,K,V)=\text{SoftMax}(QK^\intercal/\sqrt{d})$. In `w/ GHU enhance only',  we remove $\log(\mathbf{z^g})$ from Eqn.~\ref{eqn:gate_val} that suppresses history frames, \ie $G = \mathbf{z^g}$. Conversely, in `w/ GHU suppress only', we remove $\mathbf{z^g}$ from Eqn.~\ref{eqn:gate_val} that enhances history frames, \ie $G = \log(\mathbf{z^g})$. In `w/ GHU w/o position guidance', we operate on frame features before subjecting them to learned position encoding, \ie $G = \log(\mathbf{z^{\tilde{g}}}) + \mathbf{z^{\tilde{g}}}$ where $\mathbf{z^{\tilde{g}}} = q(\mathbf{h})\mathbf{E}$. We also compare with `w/ GHU per head' where G is learned separately for each cross-attention head.

Table~\ref{tab:ghu_bg_fah}a shows that our implementation of GHU significantly outperforms all other variants of GHU and cross-attention. 
We can observe that `w/o GHU' performs $1.1\%$ worse than `w/ GHU'. This is because, without explicit gating, vanilla cross-attention fails to learn attention weights for history frames that correlate with how informative history frames are to current frame prediction~(also depicted in Figure~\ref{fig:teaser}).
Moreover, the lower performances of `w/ GHU suppress only' and `w/ GHU enhance only' validate that we need to both enhance the informative history frames and suppress the uninformative ones to achieve the best performance. Without the ability to both enhance and suppress, the model may encode uninformative history frames into the latent encoding or inadequately emphasize the informative ones, leading to worse performance. 
 The performance is also lower when using history frame features without position encoding (`w/ GHU w/o position guidance'). This is because without position guidance, the model cannot assess the relative position of a particular history frame \wrt the current frame which is an important factor in deciding how informative a history frame is to current frame prediction. We also find having separate G per head (`w/ GHU per head) performs much worse than sharing G across heads due to overfitting from $N_{heads}$ times more parameters.
{\noindent \bf Impact of Background Suppression.} We compare our background suppression objective with standard cross-entropy loss~(\ie, $\gamma_a = \gamma_b = 0$) and standard focal loss(\ie, $\gamma_a = \gamma_b \neq 0$)~\cite{lin2017focal} as shown in Table~\ref{tab:ghu_bg_fah}b. First, compared to our background suppression objective, both standard cross-entropy and focal loss achieve lower accuracy. This validates that it is important to put separate emphasis on the low-confident action and background predictions to effectively differentiate action frames and closely resembling background frames. Furthermore, we find that across different combinations of $\gamma_a$ and $\gamma_b$, choosing a pair where $\gamma_a > \gamma_b$ leads to higher accuracy. Specifically, we find $\gamma_a = 0.05$ and $\gamma_b = 0.025$ to give the highest accuracy. This can be attributed to the high data imbalance. Action frames are much lower in number than background frames and therefore require a stronger emphasis than background. 

\begin{table*}[t]
\centering
\setlength{\tabcolsep}{4pt}
\resizebox{\textwidth}{!}{%
\begin{tabular}{ccc}  
\begin{tabular}{@{}l|lc@{}}
\hline
Method & mAP~(\%)    \\ \hline
Ours & \textbf{70.7} \\
w/o self-attention             & 67.7 \\
w/ cross-attention only at layer 1            & 68.6 \\
w/ disjoint history and present & 69.4\\\hline
\end{tabular}
&
\multicolumn{2}{c}{
\begin{tabular}{@{}l|c|c|cccc|c|@{}c@{}}  \hline
\multirow{3}{*}{Method}    & \multicolumn{2}{c|}{Model} & \multicolumn{5}{c|}{Inference Speed (FPS)} & \multirow{2}{*}{{ mAP(\%)}} \\ \cline{2-8}
& \multirow{2}{*}{\makecell[c]{Parameter\\Count}} & \multirow{2}{*}{GFLOPs} & \multirow{2}{*}{\makecell[c]{Optical Flow\\Computation}} & \multirow{2}{*}{\makecell[c]{RGB Feature\\Extraction}}  & \multirow{2}{*}{\makecell[c]{Flow Feature\\Extraction}}  & Model & Overall & \\
&  & &  &  &  &  & & \\ \hline
TRN~\cite{xu2021co}   & 402.9M &    1.46    & 8.1          & 70.5                 &      14.6                   & 123.3    &  8.1  & 62.1       \\
OadTR~\cite{wang2021oadtr}   & 75.8M & 2.54  & 8.1          & 70.5                 & 14.6                    &     110.0     & 8.1 & 65.2 \\
LSTR~\cite{xu2021long}(Flow-free)   & 54.2M & 6.33  & -       & 22.7                     & -                  &   99.2   & 22.7 & 63.5 \\ 
LSTR~\cite{xu2021long}   & 58.0M  &   7.53   & 8.1        & 70.5                     & 14.6                    & 91.6     & 8.1 &69.5 \\ \hline
Ours~(Flow-free) & 41.8M & 3.47 & -           &  22.7                      & -                     &  83.3        & \textbf{22.7} & 66.5 \\ 
Ours    & 45.2M  &  6.98   & 8.1         &  70.5                       & 14.6                       &    71.2    & 8.1 & \textbf{70.7} \\ \hline
\end{tabular}
}
     \\
     (a) & \multicolumn{2}{c}{(b)} \\
\end{tabular}
}
\vspace{-1em}
\caption{ (a) Ablation study for Present Decoder. (b) Efficiency comparison of \methodname~using RGB and optical flow features and our optical flow-free version with existing methods. \methodname~using RGB and optical flow has the least parameter count compared to existing methods, and higher accuracy and lower GFLOPs than the existing best method. Moreover, our flow-free version attains higher or close accuracy compared to existing methods that require RGB and optical flow features at $2.8\times$ faster inference speed.}
\label{tab:present_efficiency}
\vspace{-1.25em}
\end{table*}
{\noindent \bf Impact of Future-augmented History~(FaH).}
Table~\ref{tab:ghu_bg_fah}c shows the ablation on FaH. Since the TSN backbone is not compatible with multi-frame input, we conduct this study using RGB features from TimeSformer. The table shows that with 2s of future information incorporated into history features, we achieve the best accuracy which is $1\%$ higher than without future-augmented history~(`w/o FaH'). The accuracy is also improved with 1s of future information incorporated into history features. We further observe that the accuracy drops when future duration is much longer \eg 4s or much shorter \eg 0.5s. This shows that making a history frame aware of its future enables it to be more informative for current frame prediction. At the same time, future duration up to a certain extent~(in our case, 2s) can encode meaningful future into history frames. Much beyond that, the future changes enough to be of little use for a given history frame, while much shorter future duration may also add noise rather than information. We wish to emphasize that all future duration are bound by the frames observed so far and do not extend into inaccessible future frames.

{\noindent \bf \methodname~Present Decoder.} Table~\ref{tab:present_efficiency}a shows the ablation study on our Present Decoder by altering different aspects of the design. Unlike the original PerceiverIO~\cite{jaegle2021perceiverio}, where the output queries are independent, we model the present~(equivalent of output queries in our method) via a causal self-attention and cross-attend it with history encoding multiple times (inspired by Perceiver~\cite{jaegle2021perceiver}). We can observe in Table~\ref{tab:present_efficiency}a that treating present frames independently~(`\ie w/o self-attention') and having only one cross-attention~(`\ie w/ cross-attention only at first layer') both reduce the accuracy considerably. Unlike LSTR~\cite{xu2021long} that uses a FIFO queue with disjoint long-term and short-term memory, in our design, the sequences of history and present frames fully overlap. Table~\ref{tab:present_efficiency}a shows that having disjoint history and present frames (\ie, `w/ disjoint history and present') leads to a $1.3\%$ lower performance, further validating our design of Present Decoder and \methodname~overall.

\subsection{\methodname~Efficiency}
For online action detection setting, model efficiency is an important metric.
We compare \methodname~with existing methods \wrt parameter count, GFLOPs, and inference speed in terms of FPS as shown in Table~\ref{tab:present_efficiency}b. We first observe that \methodname~achieves the highest accuracy with the least number of model parameters compared to all existing methods. We also note that while methods like OadTR~\cite{wang2021oadtr} and TRN~\cite{xu2019temporal} are more efficient in terms of GFLOPs, their accuracy is much lower. \methodname~achieve a more favorable accuracy-efficiency trade-off with fewer GFLOPs than the existing best method LSTR~\cite{xu2021long} while obtaining a higher accuracy. All aforementioned methods require optical flow computation which is time-consuming, therefore the inference speed of these methods is governed by the optical flow computation speed of 8.1 FPS.
Meanwhile, our flow-free model obviates optical flow computation by using RGB features from TimeSformer at two different frame rates, and attains higher or close accuracy compared to existing work at $2.8\times$ faster inference speed. 
When compared with flow-free LSTR,~\methodname~achieves 3\% higher mAP, thus providing a significantly better speed-accuracy tradeoff than the existing best method.
\begin{figure}[h]
\begin{center}
    \includegraphics[width=\columnwidth]{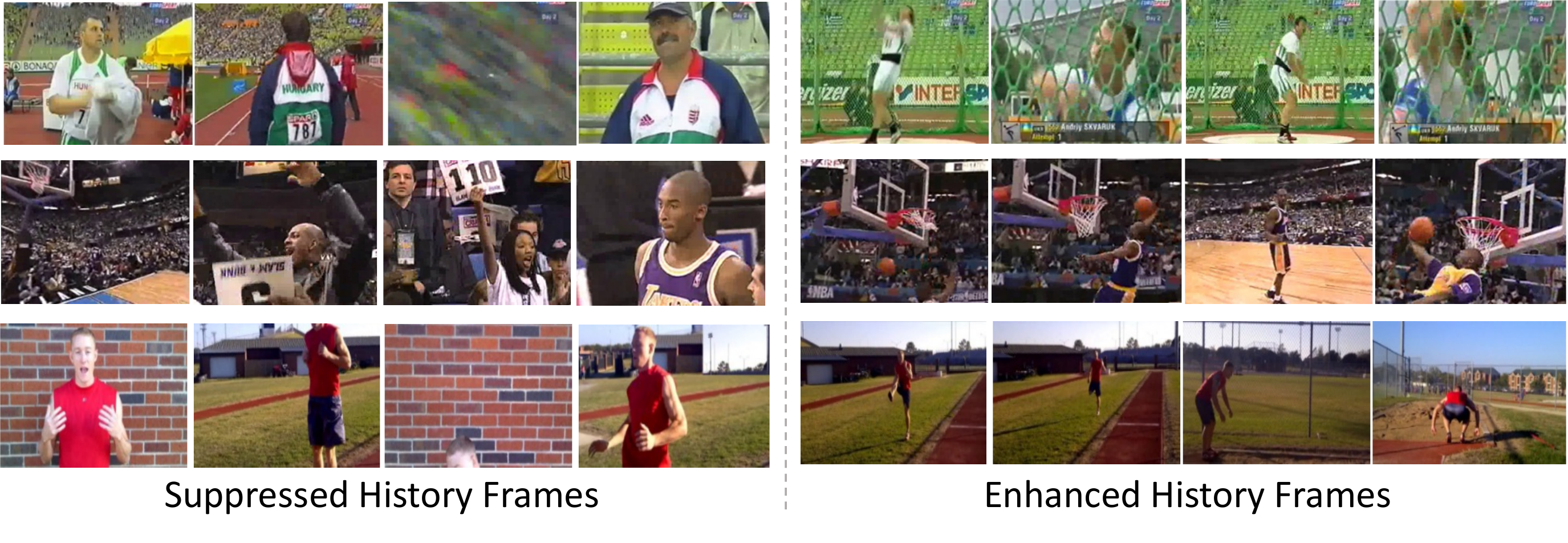}
\end{center}
\vspace{-2em}
  \caption{Examples of the most suppressed and most enhanced history frames as per the gating score learned by GHU. Frames in the same row belong to the same video. 
  }
\label{fig:ghu_vis}
\vspace{-1em}
\end{figure}
\subsection{Qualitative Evaluation}
{\noindent \bf Gated History Unit~(GHU).} We qualitatively assess the effect of GHU by visualizing examples of the most suppressed and most enhanced history frames in a streaming video when ordered as per the gating scores $G$ learned by GHU in Eqn.~\ref{eqn:gate_val}. Fig.~\ref{fig:ghu_vis} shows examples from three videos where frames in the same row belong to the same video. From the figure, we can observe that GHU learns to suppress frames that exhibit no discernible action from the $C$ action classes. The suppressed frames either have people arbitrarily moving or are uninformative background frames (\eg crowd cheering) that convey no useful information to predict action for the current frame. On the other hand, GHU learns to maximize emphasis on history frames with action from the $C$ classes and on background frames that provide meaningful context to determine the current frame action (\eg long jump athlete running toward the pit).
\begin{figure}[t]
\begin{center}
    \includegraphics[width=\linewidth]{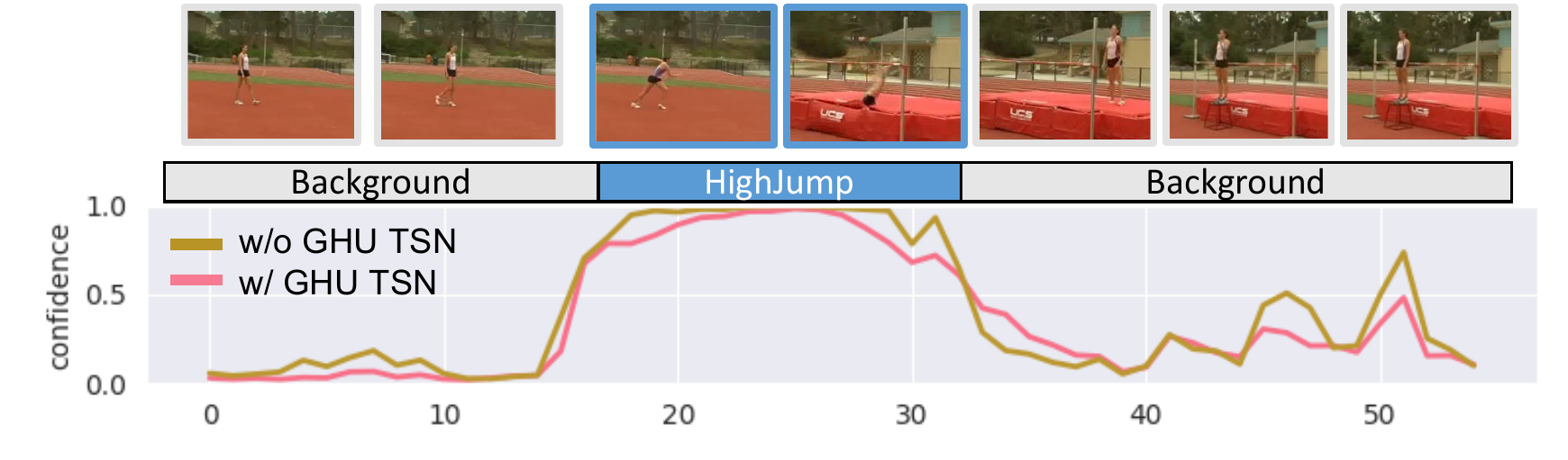}
\end{center}
\vspace{-2em}
  \caption{Visualization of \methodname's online prediction. The curves indicate the predicted confidence of the ground-truth class (\textit{High Jump}) using TSN backbone with and without GHU.
  }
  \vspace{-1.5em}
\label{fig:oad_vis}
\end{figure}

{\noindent \bf Current Frame Prediction.} We visualize \methodname's current frame prediction in Fig.~\ref{fig:oad_vis}. The confidence in the range $[0,1]$ on y-axis denotes the probability of predicting the correct action~(\ie \textit{High Jump} in Fig.~\ref{fig:oad_vis}).
We can observe that \methodname~with GHU~(red) is effective in 
reducing false positives for background frames that closely resemble action frames compared to without GHU~(orange). 
Please refer to supplementary material with visualizations highlighting more online action detection scenarios.

\section{Conclusion and Future Work}
We present \methodname~for online action detection in untrimmed streaming videos. It consists of novel designs including Gated History Unit~(GHU), Future-augmented History~(FaH), and a background suppression loss to more informatively leverage history and reduce false positives for current frame prediction. \methodname~achieves higher accuracy than all existing methods for online action detection, and is more efficient than the existing best method. Moreover, its optical flow-free variant is $2.8\times$ faster than previous methods that require both RGB and optical flow while obtaining higher or close accuracy.  

While \methodname~outperforms all existing methods, there is ample room for improvement. Although \methodname~can leverage long history, the length is still finite and may not be adequate when actions occur infrequently over long duration. It would be worthwhile to investigate ways to leverage history sequences of any length. Another challenge is slow motion action which is uncommon and can have considerably different temporal distribution, making it difficult to predict as accurately as common actions.

~\\[0pt]
\textbf{Acknowledgements}:
At Rochester Institute of Technology, Junwen Chen and Yu Kong are supported by NSF SaTC award 1949694, and the Army Research Office under grant number W911NF-21-1-0236. The views and conclusions contained in this document are those of the authors and should not be interpreted as representing the official policies, either expressed or implied, of the Army Research Office or the U.S. Government.
{\small
\bibliographystyle{ieee_fullname}
\bibliography{egbib}
}

\end{document}


\title{Supplementary: GateHUB: Gated History Unit with Background Suppression for Online Action Detection}

\author{Junwen Chen$^\star$$^\ddag$ ~~~~ Gaurav Mittal$^\star$$^\dag$ ~~~~ Ye Yu$^\dag$ ~~~~ Yu Kong$^\ddag$ ~~~~  Mei Chen$^\dag$\\
$^\dagger$Microsoft ~~~~~~~~~~~~~~~~~~~~~~~~ $^\ddag$Rochester Institute of Technology\\
{\tt\small \{gaurav.mittal, yu.ye, mei.chen\}@microsoft.com} ~~~
{\tt\small \{jc1088,yu.kong\}@rit.edu}
}
\maketitle
\thispagestyle{empty}

In this document, we provide additional analysis of our method \methodname~both quantitatively and qualitatively. We include details and analysis that were ready at the time of submission but could not be included in the paper due to the space constraints. Besides this document, we also include few videos demonstrating the task of online action detection. 

\section{Implementation Details}
\let\thefootnote\relax\footnote{$^\star$Authors with equal contribution.}
\let\thefootnote\relax\footnote{This work was done as Junwen Chen’s internship project at Microsoft.}
To extract features from TSN~\cite{wang2016temporal}, we take the average of RGB features of 6 consecutive frames at 24 FPS to represent each frame at 4 FPS. Similarly, we stack optical flow maps of 5 frames preceding each frame  along channel dimension at 24 FPS to obtain optical flow features for each frame at 4 FPS. Since TimeSformer~\cite{bertasius2021space} requires an input of $96$ RGB frames, we uniformly sample $96$ frames from the time duration set for past frames, $t_{ps}$, and future frames, $t_{f}$, for Future-augmented History~(FaH)  as input to TimeSformer and use the output corresponding to the last frame as the feature for a history frame. 


\section{Additional Quantitative Analysis}

\subsection{Future-augmented History~(FaH) on I3D} 
\begin{table}[h]
\centering
\resizebox{0.45\columnwidth}{!}{%
\begin{tabular}{lc}
\hline
Method &  mAP~(\%)    \\ \hline
WOAD~\cite{gao2020woad} & 67.1 \\
w/o FaH    & 68.1  \\ 
\multirow{1}{*}{w/ FaH}       & \textbf{69.1}  \\  \hline
\end{tabular}
}
\caption{Ablation study for Future-augmented History~(FaH) using I3D~\cite{carreira2017quo}. With FaH significantly outperforms both WOAD and without FaH. Without using FaH also outperforms WOAD showing that the other novel aspects of our method~(\ie GHU and background suppression objective) are also instrumental in improving online action detection even with I3D.}
\label{tab:supp_fah_i3d}
\end{table}
In Table 3c of the main paper, we show an ablation study on the impact of using Future-augmented History~(FaH) with TimeSformer~\cite{bertasius2021space} feature backbone. We use TimeSformer as it supports multi-frame input and is therefore compatible with FaH. To further highlight the applicability and benefit of FaH on different spatio-temporal feature backbones supporting multi-frame input, we also conduct an ablation study with I3D~\cite{carreira2017quo}. Similar to TSN~\cite{wang2016temporal}, I3D is a commonly used feature backbone for online action detection in prior art~\cite{gao2020woad}. Table~\ref{tab:supp_fah_i3d} shows the results for the ablation on FaH using I3D. We conducted the experiments on THUMOS'14. For comparison, we also provide the accuracy achieved by the existing method, WOAD~\cite{gao2020woad}, that uses I3D for THUMOS'14. From the table, we can observe that using I3D with FaH in \methodname~(`w/ FaH') significantly outperforms both WOAD and using I3D without FaH in \methodname~(`w/o FaH'). This highlights the significance of the proposed FaH module to make the history frames more informative using their \textit{future}, \ie subsequently observed frames. This, in turn, improves the history encoding and accuracy of current frame prediction. This also highlights that FaH 
can be successfully applied to improve accuracy on other spatio-temporal feature backbones that support multi-frame input. Moreover, we can observe that even without using FaH with I3D in \methodname~(`w/o FaH'), the accuracy is still $1\%$ better than WOAD. This shows that other novel aspects of \methodname, \ie using Gated History Unit~(GHU) to enhance or suppress history frames based on how informative they are to current frame prediction and using background suppression to apply separate emphasis on low-confident action and background frame predictions, are also instrumental in improving the accuracy regardless of the feature backbone.

\begin{table*}[t]
    \footnotesize
    \centering
    \setlength\tabcolsep{2.5pt}
    \resizebox{0.9\textwidth}{!}{
        \begin{tabular}{lcccccccccc}
            \hline
            \multirow{2}{*}{Method} &  \multicolumn{10}{c}{Portion of Action} \\
            \cline{2-11} 
            & \makecell{0-10\%} & \makecell{10-20\%} & \makecell{20-30\%} & \makecell{30-40\%} & \makecell{40-50\%} & \makecell{50-60\%} & \makecell{60-70\%} & \makecell{70-80\%} & \makecell{80-90\%} & \makecell{90-100\%} \\
            \hline
            
            IDN~\cite{eun2020learning} &  81.7 & 81.9 & 83.1 & 82.9 & 83.2 & 83.2 & 83.2 & 83.0 & 83.3 & 86.6 \\
            PKD~\cite{zhao2020privileged}                        & 82.1 & 83.5 & 86.1 & 87.2 & 88.3 & 88.4 & 89.0 & 88.7 & 88.9 & 87.7 \\
            OadTR~\cite{wang2021oadtr} & 81.2 & 84.9 & 87.4 & 87.7 & 88.2 & 89.9 & 88.9  & 88.8 & 87.6  & 86.7 \\
            LSTR~\cite{xu2021long}                                       & 84.4 & 85.6 & 87.2 & 87.8 & 88.8 & 89.4 & 89.6 & 89.9 & 90.0 & 90.1 \\
            \methodname~(Ours)                                      & \textbf{84.5} & \textbf{87.6} & \textbf{89.5} & \textbf{90.0} & \textbf{90.2} & \textbf{91.0} & \textbf{91.3} & \textbf{91.3} & \textbf{91.3} & \textbf{90.7} \\
            \hline
        \end{tabular}
    }
    \caption{
        Evaluation on TVSeries by dividing all action occurrences into ten equal parts (\ie \textit{portions of action}) and computing a separate mcAP(\%) for each portion of action. \methodname~outperforms all existing methods for all portions of action considered. This shows that \methodname~is more accurate in predicting for the current frame irrespective of whether it is the start, middle or end of an action occurrence.
    }
    \label{tab:supp_tvseries_stage}
\end{table*}
\subsection{Additional comparison on TSN pretrained on ActivityNet}
As shown in Table 1 of the main paper, we compare \methodname~with existing state-of-the-art~(SoTA) methods on the standard setting of using RGB and optical flow features from TSN~\cite{wang2016temporal} pretrained on Kinetics-400~\cite{carreira2017quo}. Earlier approaches~\cite{xu2019temporal} often compare with the setting of using the same TSN backbone but pretrained on ActivityNet~\cite{caba2015activitynet}. So in addition to results on TSN and TimeSformer pretrained on Kinetics in the main paper, we compare \methodname~with SoTA methods on THUMOS'14 on this setting of using RGB and optical flow features from TSN pretrained on ActivityNet. We present the results in Table~\ref{tab:supp_thumos}. From the table, we can observe that \methodname~is able to significantly outperform all existing methods by at least $3.8\%$ on this setting. Compared to TSN pretrained on Kinetics, this setting gives consistently lower accuracy for all methods. Table~\ref{tab:supp_tvseries} further shows the results on TVSeries. We can again observe that \methodname~is able to outperform all existing methods. This further validates that \methodname~can outperform all existing methods on multiple benchmark datasets using multiple different input feature representations.
 \begin{table}[h]
\centering
\resizebox{0.5\columnwidth}{!}{%
\begin{tabular}{lc}
\hline
\multirow{1}{*}{Method}  & mAP~(\%)  \\ \hline  
      CDC~\cite{shou2017cdc}     & 44.4       \\
      RED~\cite{gao2017red}      & 45.3      \\
TRN~\cite{xu2019temporal}   &     47.2         \\
FATS~\cite{kim2021temporally} & 51.6 \\
IDN~\cite{eun2020learning} & 50.0 \\
LAP~\cite{qu2020lap} &  53.3 \\
TFN~\cite{eun2021temporal} & 55.7 \\
OadTR~\cite{wang2021oadtr} &   58.3    \\
LSTR~\cite{xu2021long}  &     65.3      \\
\methodname~(Ours)  &      \textbf{69.1}        \\ \hline
\end{tabular}%
}
\caption{Online action detection results on THUMOS'14 comparing \methodname~with SoTA methods on mAP~(\%) when the RGB and optical flow-based features are extracted from TSN pretrained on ActivityNet. We can see that \methodname~significantly outperforms all existing methods.}
\label{tab:supp_thumos}
\end{table}
 \begin{table}[h]
\centering
\resizebox{0.5\columnwidth}{!}{%
\begin{tabular}{lc}
\hline
\multirow{1}{*}{Method}  & mcAP~(\%)  \\ \hline  
      RED~\cite{gao2017red}      & 79.2   \\
FATS~\cite{kim2021temporally} & 81.7 \\
TRN~\cite{xu2019temporal}   &     83.7         \\
IDN~\cite{eun2020learning} & 84.7 \\
TFN~\cite{eun2021temporal} & 85.0 \\
LAP~\cite{qu2020lap} &  85.3 \\
OadTR~\cite{wang2021oadtr} &   85.4   \\
LSTR~\cite{xu2021long}  &     88.1      \\
\methodname~(Ours)  &      \textbf{88.4}        \\ \hline
\end{tabular}%
}
\caption{Online action detection results on TVSeries comparing \methodname~with SoTA methods on mcAP~(\%) when the RGB and optical flow-based features are extracted from TSN pretrained on Activity Net. We can see that \methodname~outperforms all existing methods.}
\label{tab:supp_tvseries}
\end{table}

\subsection{Evaluation on TVSeries for different portions of action} 
Following prior art~\cite{de2016online, xu2019temporal, wang2021oadtr, xu2021long}, we also evaluate the accuracy of online action detection on TVSeries when only a certain portion of the action occurrences is considered. The objective of this evaluation is to assess how well a method performs at different stages of an ongoing action. Following prior art, we divide each action occurrence into ten equal parts~(\ie \textit{portions of action}). We then compute a separate mcAP for each portion of action over all action occurrences. For example, mcAP for $20-30\%$ \textit{portion of action} refers to mcAP computed using only frames of an action occurrence lying between $20\%$ and $30\%$ of the total duration of that action occurrence. We tabulate the results across all portions of action in Table~\ref{tab:supp_tvseries_stage}. From the table, we can observe that our method outperforms all existing methods for all the different portions of action considered. This shows that irrespective of whether it is the start, middle or end of an action occurrence, \methodname~is able to predict the action for the current frame more accurately than all existing methods.


\subsection{Effect of History and Present duration}

\begin{table}[h]
\centering
\resizebox{0.7\columnwidth}{!}{%
\begin{tabular}{c|cccc}
\hline
 \multirow{2}{*}{\makecell[c]{History\\Duration~(s)}} & \multicolumn{4}{c}{Present Duration~(s)}    \\ \cline{2-5}
 & 1 & 2 & 4 & 8  \\ \hline
  64   & 69.2 & 68.4 & 67.9 & 67.8    \\ 
  128   & 68.3 & 67.9 & 68.5 & 65.8    \\ 
  256   & 69.3 & \textbf{70.7} & 69.9 & 67.4   \\ 
  512   & 69.1 & 68.6 & 68.4 & 68.0    \\  \hline
  
  \end{tabular}
}
\caption{Ablation study showing mAP~(\%) for different durations (in seconds) of history and present frames sampled at 4 FPS using RGB and optical flow features from TSN on THUMOS'14.}
\label{tab:supp_history_present}
\end{table}

We show an analysis where we experiment with different durations of history and present frames in \methodname~in Table~\ref{tab:supp_history_present}. We test on THUMOS'14 using RGB and optical flow features from TSN~\cite{wang2016temporal}. For each setting, the frames are sampled at 4 FPS. We consider a duration for history ranging from 64s to 512s and for present ranging from 1s to 8s~(duration doubling for each subsequent setting). From the table, we can observe that we can get the best accuracy using history and present of duration 256s and 2s respectively. We can also observe that the model gets the worst performance when the duration of present is 8s for any given duration of history. This suggests that the present should constitute a very small set of most recently observed frames preceding the current frame. This allows to effectively model the most immediate observable context around the current frame which is important for accurate prediction for the current frame.

\subsection{Action Anticipation Result}
Following LSTR~\cite{xu2021long}, we evaluate GateHUB on action anticipation task. Specifically, we anticipate the future up to 2 seconds at 4FPS by adding 8 learnable tokens to the most recent history frames $[h_t]_{t=-t_{pr}-1}^0$. Table~\ref{tab:supp_anticipate} shows that GateHUB significantly outperforms the existing methods by $4.1\%$ and $1.2\%$ on THUMOS and TVSeries respectively, using the ActivityNet pretrained features. 
\begin{table}[h]
\centering
\resizebox{0.7\columnwidth}{!}{%
\begin{tabular}{lcc}
\hline
\multirow{1}{*}{Method} &mAP~(\%) & mcAP~(\%)  \\ \hline  
      RED~\cite{gao2017red}   &37.5   & 75.1   \\
TRN~\cite{xu2019temporal}   & 38.9  &  75.7         \\
TTM~\cite{wang2021ttpp} & 40.9 & 77.9\\
LAP~\cite{qu2020lap} &  42.6  &78.7 \\
OadTR~\cite{wang2021oadtr} &45.9  & 77.8   \\
LSTR~\cite{xu2021long}  &  50.1 &  80.8      \\
\methodname~(Ours)  &  \textbf{54.2}     &\textbf{82.0}        \\ \hline
\end{tabular}%
}
\caption{Results of online action anticipation using ActivityNet features on THUMOS'14 and TVSeries in terms of mAP and mcAP, respectively.}
\label{tab:supp_anticipate}
\end{table}

\subsection{Analysis of gating scores $G$ vs $Q_iK_i^T/\sqrt{dk}$}
\begin{figure}[h]
\vspace{-1em}
\begin{center}
    \includegraphics[width=\linewidth]{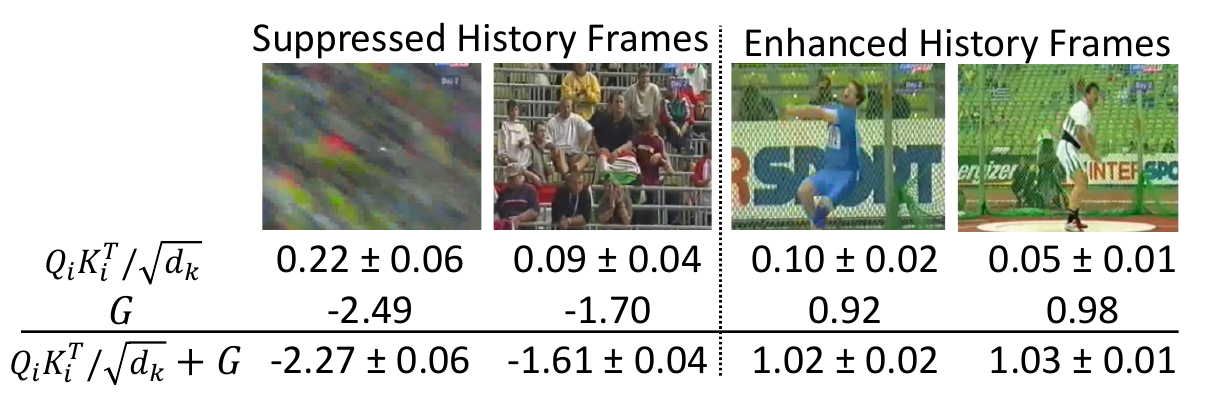}
\end{center}
\vspace{-2em}
  \caption{Analysis of gating scores $G$ vs $Q_iK_i^T/\sqrt{dk}$ for the some of the most suppressed and enhanced history frames
  }
  \vspace{-1em}
\label{fig:g_analysis}
\end{figure}

We compare the value of $G$ with $Q_iK_i^T/\sqrt{dk}$ to assess whether the gating scores $G$ are indeed able to calibrate the attention weights. Statistically, on obtaining $G$ and $Q_iK_i^T/\sqrt{dk}$ across all history frames, we find that G lies in $[-9.5, 1.0)$ and $Q_iK_i^T/\sqrt{dk}$ lies in $[-0.1, 0.6]$. So, G is large/small enough to change relative order of attention scores. Further, Fig~\ref{fig:g_analysis} further provides G and range of $Q_iK_i^T/\sqrt{dk}$ and $Q_iK_i^T/\sqrt{dk} + G$~(Eqn 3, main paper) for some frames in Fig 3 of main paper. We can see that G is able to calibrate $Q_iK_i^T/\sqrt{dk}$ so that informative and uninformative history frames are correctly enhanced and suppressed respectively. 

\begin{figure*}[t]
\begin{center}
\includegraphics[width=\textwidth]{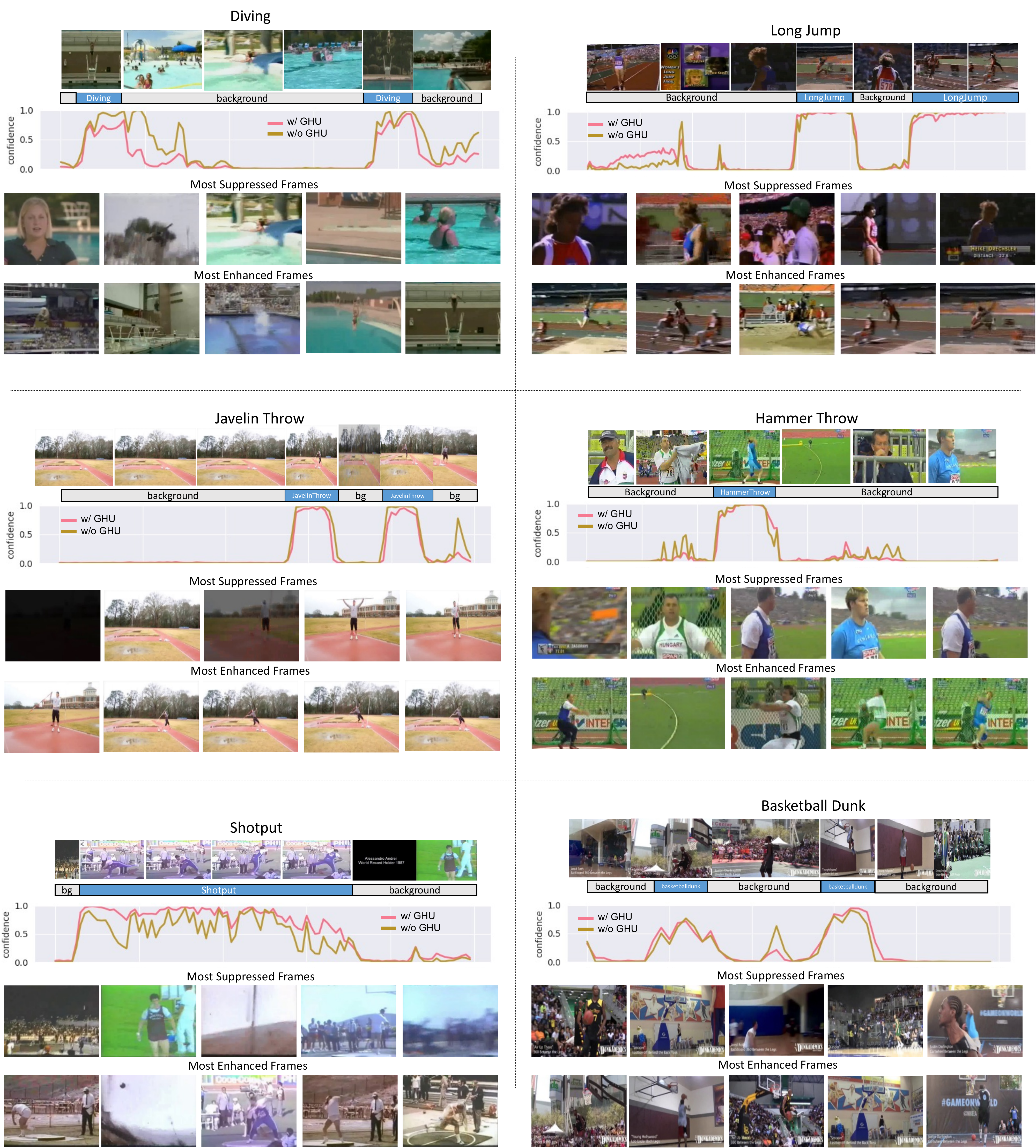}
\end{center}
\vspace{-1em}
\caption{Visualizing the current frame prediction for six videos from THUMOS'14 (separated by dotted lines). For each video, the first row shows the video frames, then ground truth~(blue denoting action occurrence) followed by the plot for current frame prediction comparing \methodname~with GHU~(`w/ GHU')~(red) and without GHU~(`w/o GHU')~(brown). Below the plots for each video, we also highlight examples of the most suppressed and the most enhanced frames in the corresponding video as ordered based on the gating score G in \methodname~with GHU. `w/ GHU' is able to reduce false positives observed in `w/o GHU' where background frames closely resemble action frames.} 
\label{fig:supp_vis}
\end{figure*}

\begin{figure*}[t]
\begin{center}
\includegraphics[width=\textwidth]{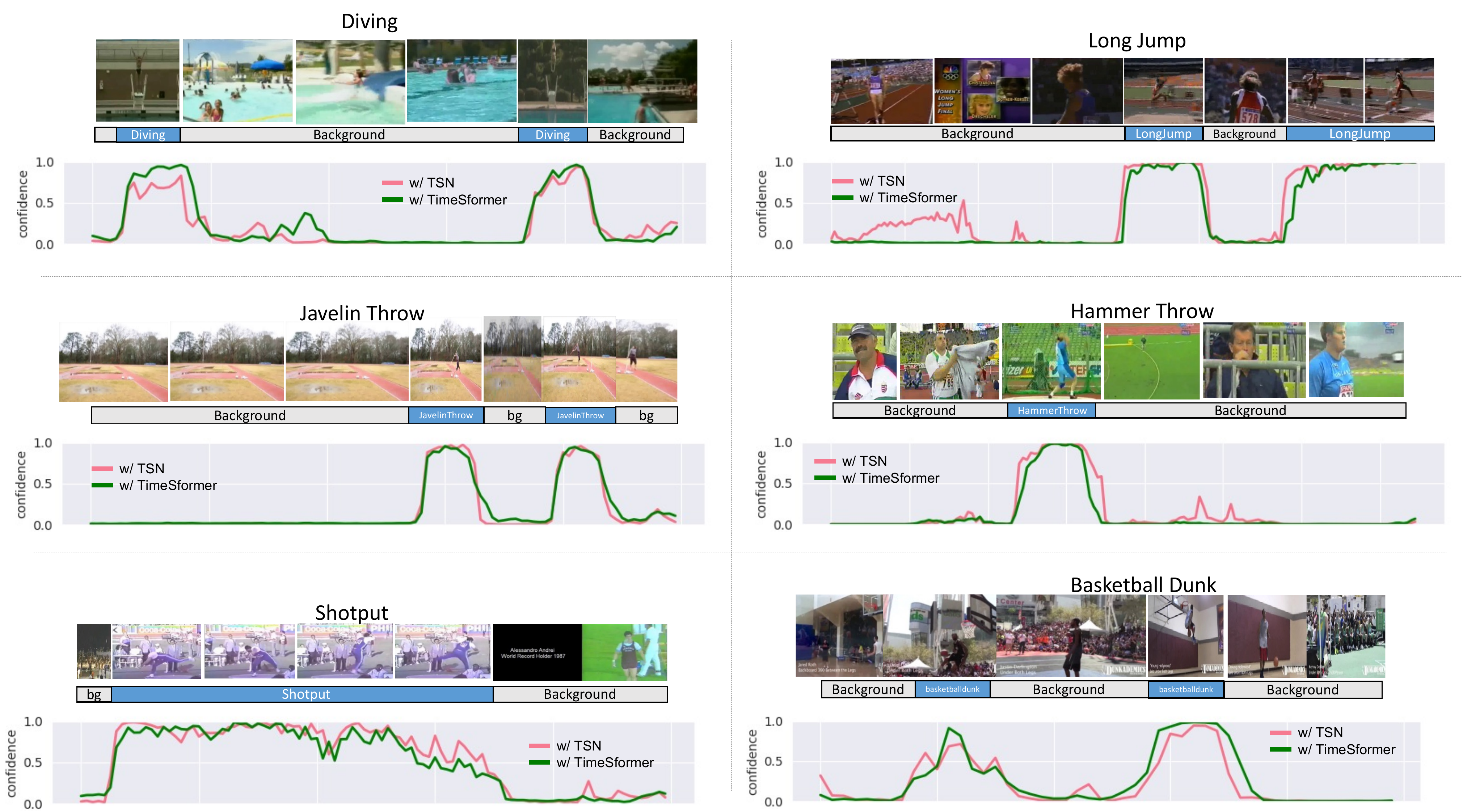}
\end{center}
\caption{Visualizing the current frame prediction for six videos from THUMOS'14 (separated by dotted lines). For each video, the first row shows the video frames, then ground truth~(blue denoting action occurrence) followed by the plot for current frame prediction comparing \methodname~using RGB features from TSN~(red) and TimeSformer~(green).} 
\label{fig:supp_vis_tsntfk}
\end{figure*}

\section{Additional Qualitative Analysis}
We provide additional qualitative assessment by visualizing \methodname's current frame prediction with and without GHU in Fig.~\ref{fig:supp_vis}. 
There are six video examples from THUMOS’14. For each example, we show the video frames at the top, then the ground truth~(blue denoting the action occurrences), followed by current frame predictions 
using \methodname~with GHU~(red) and without GHU~(brown) 
where the confidence in the range $[0,1]$ on y-axis denotes the probability of predicting the correct action. At the bottom of each example, we present the most suppressed and the most enhanced frames determined by GHU. From the visualization, we can observe that when using GHU (red), our method is able to significantly reduce false positives for the background frames~(particularly that closely resemble action frames such as the frames closely following the second \textit{Diving} and \textit{Javelin Throw} action occurrence in Fig.~\ref{fig:supp_vis}). At the same time, without GHU (brown), the model is prone to high number of false positives~(as can be been in the \textit{Diving} example after the first action occurrence for frames showing swimming pool). In addition to reducing false positives, using GHU is also able to boost the confidence of true positives~(as can been seen from \textit{Shotput} example in Fig.~\ref{fig:supp_vis}).

Below the visualization of current frame prediction for each video in Fig.~\ref{fig:supp_vis}, we also visualize examples of the \textit{most suppressed} and the \textit{most enhanced} history frames in that video when ordered as per the gating scores G learned by GHU in `w/ GHU' as per Eqn.~2 (main paper). From all the examples, we can observe that a significant number of the \textit{most suppressed frames} contain athletes as they are walking in the field either to begin preparing for the action, leave after finishing the action or stopping to respond to the interviewer. In all these scenarios, we cannot draw any meaningful context about what and when the action will begin or end. As a result, GHU helps to suppress such history frames that are highly uninformative for current frame prediction. At the same time, a significant number of \textit{most enhanced frames} are the frames where either the action is in progress or the athlete is close to commencing the action. Both these scenarios provide informative context in inferring what and when the action will take place. As as result, GHU enhances these frames that are highly informative for current frame prediction. We can also observe that occasionally few informative frames~(such as the second and fourth frame in the `most suppressed frames' for \textit{Basketball Dunk}) get suppressed by GHU. This is likely due to the action being far-away in the frame making it difficult for the model to accurately assess the informative-ness of the frame. Spatially localizing small and far-away objects and their corresponding motion is still an open challenge. Capturing more fine-grained higher resolution features could potentially mitigate the problem. 

In Fig.~\ref{fig:supp_vis_tsntfk}, we further visualize and compare \methodname's current frame prediction using RGB features from TSN~(red) and TimeSformer~(green). We can observe that \methodname~using RGB features from TimeSformer~(green) is more effective in reducing false positives while improving the confidence score for true positives~(as can be seen from false positive reduction in \textit{Long Jump} and true positive enhancement in \textit{Basketball Dunk} in Fig.~\ref{fig:supp_vis_tsntfk}). This is potentially due to the comparatively limited feature representation capacity of the TSN feature backbone to extract sufficiently discriminative features when the frames include slow motion, motion blur, or small/far-away objects. In comparison, using the Timesformer feature backbone considerably mitigates false positives~(green). This shows that with stronger feature representation, \methodname~can be more effective in reducing false positives while increasing true positives thereby improving current frame prediction. Also worth noting is that the confidence to correctly predict the \textit{Shotput} action reduces for all methods in both Fig.~\ref{fig:supp_vis} and Fig.~\ref{fig:supp_vis_tsntfk} toward the end of the action occurrence. Similar to background suppression, we can explore putting separate emphasis on accurately predicting the frames near the boundary to mitigate such low-confident near-boundary predictions.



{\small
\bibliographystyle{ieee_fullname}
\bibliography{egbib}
}